\documentclass[10pt,twocolumn,letterpaper]{article}

\usepackage[pagenumbers]{cvpr} 

\usepackage{graphicx}
\usepackage{amsmath}
\usepackage{amssymb}
\usepackage{booktabs}
\usepackage{color}
\usepackage{nicefrac}

\usepackage{etoolbox}
\makeatletter
\pretocmd{\chapter}{\addtocontents{toc}{\protect\addvspace{15\p@}}}{}{}
\pretocmd{\section}{\addtocontents{toc}{\protect\addvspace{5\p@}}}{}{}
\pretocmd{\subsection}{\addtocontents{toc}{\protect\addvspace{3\p@}}}{}{}
\makeatother
\usepackage{titletoc}

\usepackage{xpatch}
\usepackage{array}
\usepackage{multirow}
\usepackage{graphicx}
\usepackage{xcolor,colortbl}
\usepackage{pifont}
\usepackage{enumitem}
\usepackage{xcolor}
\PassOptionsToPackage{dvipsnames}{xcolor}

\definecolor{Gray}{gray}{0.9}

\definecolor{indian red}{RGB}{205,92,92}

\usepackage{comment}
\usepackage[accsupp]{axessibility}

\definecolor{mycitecolor}{rgb}{0, 0.4, 0.7}
\usepackage[pagebackref,breaklinks,colorlinks,bookmarks,citecolor=mycitecolor]{hyperref}

\usepackage[capitalize]{cleveref}
\crefname{section}{Sec.}{Secs.}
\Crefname{section}{Section}{Sections}
\Crefname{table}{Table}{Tables}
\crefname{table}{Tab.}{Tabs.}

\def\cvprPaperID{1655} 
\def\confName{CVPR}
\def\confYear{2023}

\begin{document}

\title{
Lane2Seq: Towards Unified Lane Detection via Sequence Generation
}

\author{Kunyang Zhou \\
Southeast University\\
{\tt\small kunyangzhou@seu.edu.cn}}

\maketitle


\begin{abstract}
	In this paper, we present a novel sequence generation-based framework for lane detection, called Lane2Seq. It unifies various lane detection formats by casting lane detection as a sequence generation task. 
	This is different from previous lane detection methods, which depend on well-designed task-specific head networks and corresponding loss functions. Lane2Seq only adopts a plain transformer-based encoder-decoder architecture with a simple cross-entropy loss. 
	Additionally, we propose a new multi-format model tuning based on reinforcement learning to incorporate the task-specific knowledge into Lane2Seq. Experimental results demonstrate that such a simple sequence generation paradigm not only unifies lane detection but also achieves competitive performance on benchmarks. 
	For example, Lane2Seq gets 97.95\% and 97.42\% F1 score on Tusimple and LLAMAS datasets, establishing a new state-of-the-art result for two benchmarks.
\end{abstract}

\section{Introduction}
\begin{figure}[t!]
  \centering
  \includegraphics[width=\linewidth]{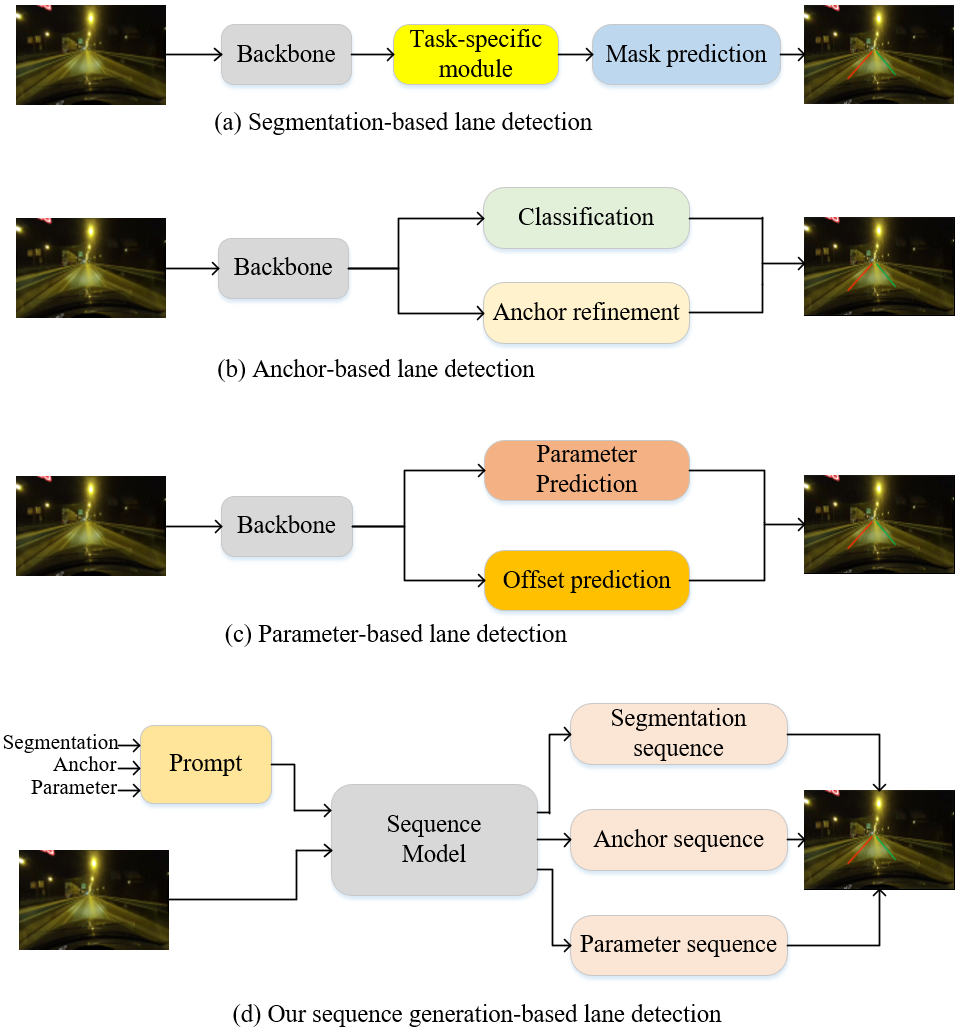}
     \caption{Comparison of different lane detection frameworks.}
  \label{fig:short1}
  \vspace{-0.5cm}
  \end{figure}

  Lane detection is a fundamental task in computer vision~\cite{qin2020ultra,tabelini2021keep,wang2018lanenet}. It aims to predict the location of the lane in a given image. Lane detection plays a crucial role in 
many applications, such as adaptive cruise control and lane keeping. Existing lane detection methods generally adopt a divide-and-conquer strategy, which decomposes the lane detection into multiple subtasks. Each subtask 
is accomplished by a task-specific head network. For instance, as shown in Fig.~\ref{fig:short1}, segmentation-based methods~\cite{zheng2021resa,pan2018spatial} adopt a head network along with a task-specific module, such as the message-passing module~\cite{pan2018spatial}, to predict per-pixel masks. Anchor-based methods~\cite{tabelini2021keep,honda2023clrernet}
utilize a classification head network for distinguishing lane instances and an anchor refinement network for regressing accurate lanes. Parameter-based methods~\cite{liu2021end,fan2019spinnet} use a network to predict the parameters of lanes and a vertical offset prediction network to locate the start points of lanes.

Although divide-and-conquer strategy has been proved an effectively way to address the certain subtask in the existing methods, there still exist several limitations. (1) Each subtask needs a customized task-specific head network, resulting in a complicated lane detection model. 
(2) Each task-specific head requires one or more loss functions, e.g., cross-entropy loss and Line-IOU loss~\cite{honda2023clrernet}, bringing out extra hyper-parameters.

In this paper, we present a novel sequence generation-based lane detecton (Lane2Seq) framework to tackle the aforementioned issues. By formulating the lane detection as a sequence generation task, Lane2Seq gets rid of customized head networks and task-specific loss function. 
It is based on the intuition that if the detection model knows where the target lane is, model can be simply teached how to read the location of lane out, instead of designing additional classification head or regression head by the divide-and-conquer strategy.

Therefore, we convert the outputs of different lane detection formats, i.e., segmentation-based, anchor-based, and parameter-based, into a sequence of discrete tokens and the model learns to generate this sequence token-by-token. As shown in Fig.~\ref{fig:short1} (d), to achieve a specific lane detection format, Lane2Seq uses a prompt to specify the detection format and 
the generated sequence adapts to the prompt so the model can produce format-specific output. By the format-specific prompt, Lane2Seq unifies different lane detection formats into a model.

While Lane2Seq does not contain task-specific components, task-specific knowledge contained in these components can help the model learn the features of lanes better. We propose a Multi-Format model tuning method based on Reinforcement Learning (MFRL) to incorporate the task-specific knowledge into the model without changing model's architecture. Inspired by Task-Reward~\cite{pinto2023tuning}, MFRL takes the evaluation metrics,
which naturally integrates task-specific knowledge, as the reward and tunes Lane2Seq using REINFORCE~\cite{williams1992simple} algorithm. However, evaluation metrics like F1 score cannot be used as the reward directly due to undecomposable as a sum of per-example rewards. In this paper, we propose three new evaluation metric-based rewards for segmentation, anchor, and parameter format, based on their task-specific knowledge.

Experimental results demonstrate that our Lane2Seq achieves competitive performance on three public datasets, Tusimple, CULane, and LLAMAS. For instance, Lane2Seq with ViT-Base encoder obtains 97.95\% and 97.42\% F1 score on Tusimple and LLAMAS, setting a new state-of-the-art result for two datasets.
It should be noted that all existing lane detection methods heavily rely on well-designed task-specific head networks and the corresponding complicated loss functions. Instead, our Lane2Seq utilizes a plain transformer-based encoder-decoder architecture with a simple cross-entropy loss.

The main contributions of this paper are as follows.
\begin{itemize}
  \item We propose a sequence generation-based method for lane detection, which casts the lane detection as a sequence generation task. To the best of our knowledge, we are the first to unify the lane detection through the sequence generation, which offers a new perspective on lane detection.
  \item We present a novel reinforcement learning-based multi-format model tuning, including three new evaluation metric-based reward functions, to incorporate the task-specific knowledge into the model.
  \item Experimental results show that our method achieves competitive performance on lane detection benchmarks. Remarkably, we establish a new state-of-the-art result on Tusimple and LLAMAS.
  \end{itemize}

\section{Related Work}
\textbf{Lane detection}. Existing lane detection methods can be divided into three categories based on the representation of the lane: 
  segmentation-based method, anchor-based method and parameter-based method. Segmentation-based methods~\cite{pan2018spatial,zhang2023houghlanenet,zheng2021resa} consider lane detection as a semantic segmentation task
  and performs pixel-wise prediction. SCNN~\cite{pan2018spatial} enhances the visual evidence by a message-passing module, which can capture spatial dependency for lanes.
  Anchor-based methods~\cite{li2019line,tabelini2021keep,qin2020ultra,zhou2023end} predict the accurate lanes by refining the predefined lane anchors. 
  UFLD~\cite{qin2020ultra} proposes a novel row anchor-based approach to detect lanes.
  Different from the segmentation-based and anchor-based methods, parameter-based methods~\cite{tabelini2021polylanenet,liu2021end,fan2019spinnet} regard the lane detection as the parametric modeling and regress the parameters of the lane.
  PolyLaneNet~\cite{tabelini2021polylanenet} models a lane curve as a polynomial and regresses the parameters of the polynomial. In this paper, we treat the segmentation-based, anchor-based methods, and parameter-based metohds uniformly as the sequence generation task, getting rid of the 
  complicate structure and task-specific modules, e.g., head network. It only adopts the cross-entropy loss and plain transformer architecture.
  
  \begin{figure*}[!t]
	\begin{center}
	\includegraphics[height = 8cm,width=15cm]{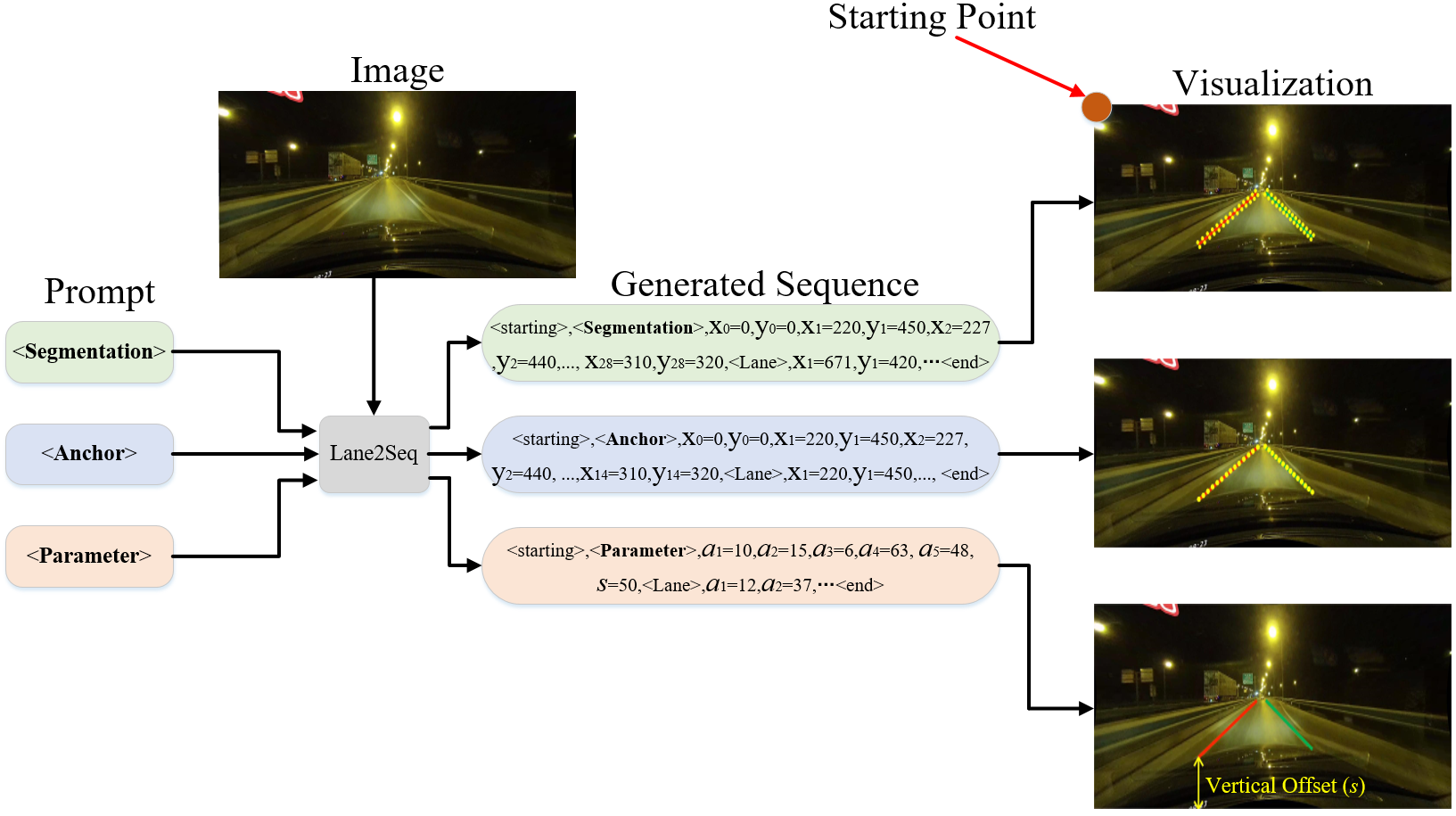}
	\end{center}
	\vspace{-0.4cm}
	   \caption{Inference pipeline of Lane2Seq. The model perceives input image, prompt and generates format-specific tokens, which can be detokenized into required detection format for visualization.}
	\vspace{-0.4cm}
	\label{fig:short_2}
	\end{figure*}
  
  \textbf{Sequence generation for vision tasks}. Recently, sequence-to-sequence (seq2seq) method used in the natural language processing (NLP) has been applied for vision tasks. Seq2seq utilizes a basic transformer encoder and decoder architecture and 
  accomplishes the task by making sequence predictions, rather than designing a model tailored to the vision task. Pix2seq~\cite{chen2021pix2seq} is the pioneering work to cast the object detection as the sequence generation task. It shows that object can be detected well without any 
  task-specific modules like label assignment. Besides object detection, seq2seq has been extended to other vision tasks such as instance segmentation~\cite{chen2022unified}, keypoint detection~\cite{chen2022unified}, text spotting~\cite{kil2023towards,peng2022spts} and object tracking~\cite{chen2023seqtrack}.
  UniTAB~\cite{yang2022unitab} adopts the task prompt~\cite{raffel2020exploring} to perform multi-task learning. Unified-IO~\cite{lu2022unified} jointly trains various vision tasks by setting the unified input/output formats for all tasks.
  Besides, seq2seq becomes increasingly popular in the multi-modal model. Text-to-image models like DALL-E~\cite{ramesh2021zero} and vision-language models like Flamingo~\cite{alayrac2022flamingo} all use seq2seq to unify the multi-modal tasks.
  Nevertheless, how to perform the seq2seq in the lane detection to unify different detection formats is still unexplored. 
  
  Lane2Seq has a similar spirit with the Pix2seq and its successors~\cite{chen2022unified,kil2023towards,peng2022spts,chen2023seqtrack}. All of them view the vision tasks as the sequence generation.
  The main difference between these methods and Lane2Seq is the sequence format. Previous works use coordinates and category to construct the sequence, while Lane2Seq also adopts the parameter of the lane to construct the sequence.
  
  \textbf{Reinforcement learning in the computer vision}. Many previously researches introduce the reinforcement learning for the vision tasks, such as object detection~\cite{mathe2016reinforcement}, object tracking~\cite{luo2018end}, image segmentation~\cite{liao2020iteratively} and lane detection~\cite{zhao2020deep}. 
  They generally focus on learning different parts of an image and refine the outputs iteratively. DQLL~\cite{zhao2020deep} localizes the lanes as a group of lanemarks and refine the location of lanemarks with the deep Q-learning. 
  Recently, Task Reward~\cite{pinto2023tuning} adopts a novel task risk-based reinforcement learning to tune the vision model without changing the model architecture. Reward functions in Task-Reward are designed for object detection, instance segmentation, colorization, and image captioning but not for lane detection. 
  Our MFRL is the first attempt to design effective evaluation metric-based reward functions for different lane detection formats.
  
  \section{Method}
  In this section, we present the proposed sequence generation-based lane detection method, named Lane2Seq, in detail. The overall pipeline of Lane2Seq is illustrated in Fig.~\ref{fig:short_2}. Sec~\ref{sec:interface} depicts the unified interface for lane detection, Sec~\ref{sec:architecture} details the model architecture and objective function,
  and Sec~\ref{sec:tuning} introduces the proposed multi-format model tuning based on reinforcement learning.

  \subsection{Unified Interface for Lane Detection}
  \label{sec:interface}
  As presented in Figure~\ref{fig:short1}, lane detection formats in the existing method are diverse and have been formulated quite differently. Considering differences in the form of outputs, customized models with specialized head networks and loss functions are designed for different lane detection formats.
  
  To integrate different lane detection formats into a model, we propose an unified sequence interface for lane detection, where both format transcription like segmentation and outputs are treated as sequences of discrete tokens. As shown in Fig.~\ref{fig:short_2}, the generated sequence is composed of four parts, a starting token \textless{starting}\textgreater{},
  a format transcription token like \textless{Segmentation}\textgreater{}, a format-specific sequence, and an ending token \textless{end}\textgreater{}. The format-specific sequence for three detection formats can be constructed as follows.
  
  \textit{Segmentation} sequence. Instead of performing pixel-wise mask prediction, we predict the polygon~\cite{castrejon2017annotating} corresponding to the mask as a sequence of coordinates conditioned on a given lane instance. Then, we convert the polygon into a sequence by quantizing the coordinates of its points
  into discrete tokens. Specifically, $x,y$ coordinates of a point are normalized to the width and height of the image, and then quantized to $\left[1,n_{bins}\right]$, where $n_{bins}$ is the size of the vocabulary. Vocabulary will be described subsequently. A polygon sequence can be expressed as [$x_{1}, y_{1},x_{2}, y_{2},...,x_{28}, y_{28}$,\textless{Lane}\textgreater{}], where \textless{Lane}\textgreater{} is the category token.
  If there are multiple lanes in an image, we concatenate the all polygon sequences. The segmentation sequence consists of a starting point and all polygon sequences, where starting point is the left-top point of the image $x_{0}=0,y_{0}=0$.
  
  \textit{Anchor} sequence. Since the essence of anchor-based methods is regressing the location of keypoints of lane anchors, we treat the anchor prediction as the keypoint sequence generation. Specifically, the keypoint sequence of a lane can be expressed as [$x_{1}, y_{1},x_{2}, y_{2},...,x_{14}, y_{14}$,\textless{Lane}\textgreater{}].
  The normalization and quantization of keypoint coordinates are same to that of polygon point. The anchor sequence contains a starting point and all keypoint sequences, where starting point is still the left-top point of the image.
  
  \textit{Parameter} sequence. the sequence of the lane includes two parts: the parameter of polynomial function and vertical offsets. We set the polynomial degree as 5th. Specifically, the parameter sequence of a lane can be represented as [$a_{1}, a_{2}, a_{3}, a_{4}, a_{5}, s$, \textless{Lane}\textgreater{}], where $s$ is the vertical offset.
  $s$ is normalized to the height of the image and quantized to $\left[1,n_{bins}\right]$. We use sigmoid to normalize $a_{i}$ by $a_{i}=Sigmoid(a_{i})$. Then $a_{i}$ is quantized to $\left[1,n_{bins}\right]$. Parameter sequence does not require the starting point.
  
  \textbf{Vocabulary}. We use a shared vocabulary for all formats and each integer between $\left[1,n_{bins}\right]$ can be regarded as a word in the vocabulary. Each word in the vocabulary corresponds to a learnable embedding. 
  
  \subsection{Unified Architecture and Objective Function}
  \label{sec:architecture}
  \begin{figure}[t!]
	\centering
	\includegraphics[width=8cm]{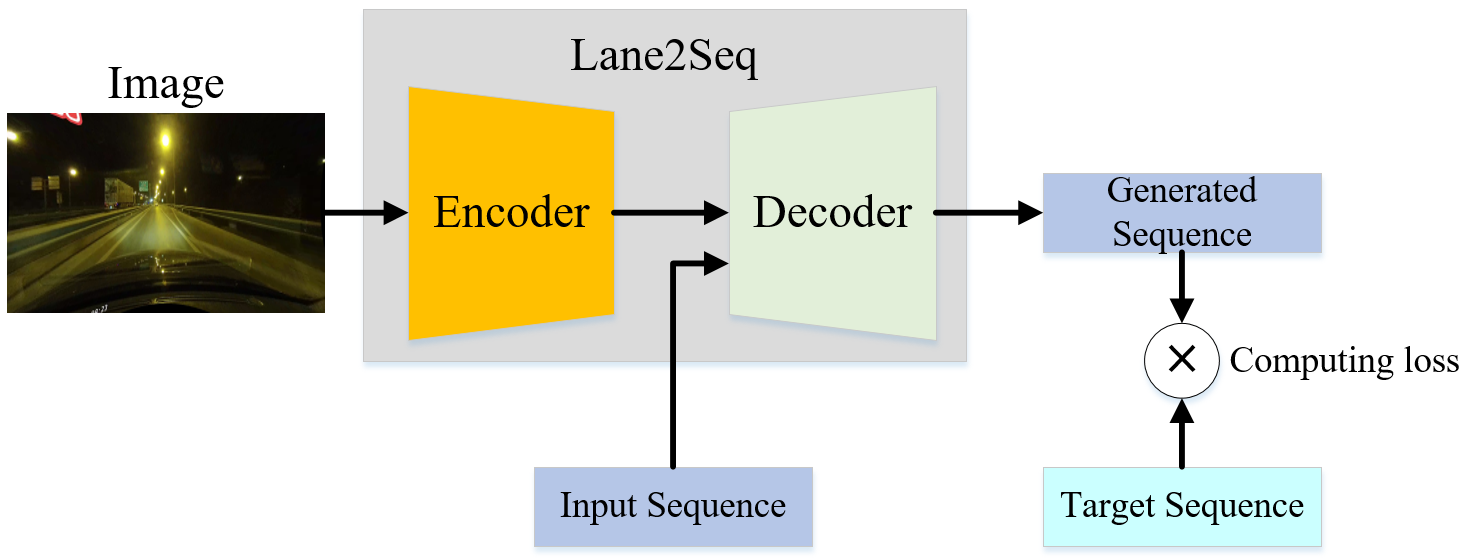}
	\caption{Training pipeline of Lane2Seq. The input sequence can be the segmentation sequence or anchor sequence or parameter sequence.}
	\vspace{-0.5cm}
	\label{fig:short3}
	\end{figure} 
  We follow~\cite{chen2023seqtrack} and use a transformer-based encoder-decoder architecture to deal with image input and sequence output flexibly. As shown in Fig.~\ref{fig:short3}, the image encoder takes the pixels as the input and outputs the corresponding image representations. We utilize the Vision Transformer (ViT)~\cite{dosovitskiy2020image} to instantiate the image encoder.
  We adopt a transformer-based sequence decoder, which is widely used in the language modeling~\cite{radford2018improving,raffel2020exploring}, to generate the output sequence. The decoder generates one token at a time based on the preceding tokens and the image representations. Sequence decoder removes the well-designed task-specific heads for different detection formats.
  
  \textbf{Training Pipeline}. We first tokenize annotation of each format into the corresponding sequence. Then, we construct a training batch using images and sequences from all detection formats. Finally, we compute loss between generated sequence and target sequence for each format. For a certain detection format, the input sequence is [\textless{starting}\textgreater{}, format transcription token, format-specific sequence] and the target sequence is [format transcription token, format-specific sequence, \textless{end}\textgreater{}]. For example, in the anchor format, 
  the input sequence is [\textless{starting}\textgreater{}, \textless{Anchor}\textgreater{}, $x_{0}, y_{0},x_{1}, y_{1}$,..., \textless{Lane}\textgreater{}, \textless{end}\textgreater{}] and the target sequence is [\textless{Anchor}\textgreater{}, $x_{0}, y_{0},x_{1}, y_{1}$,..., \textless{Lane}\textgreater{}, \textless{end}\textgreater{}]. The starting token \textless{starting}\textgreater{} and ending token \textless{end}\textgreater{} are learnable embeddings, which tell the decoder when to begin and end the sequence generation. The sequence decoder perceives the image representations and input sequence and reconstructs the target sequence. 
  
  \textbf{Objective Function}. Similar to Pix2seq~\cite{chen2021pix2seq}, we train the Lane2Seq with a simple cross-entropy loss. At each time stamp $j$, Lane2Seq aims to maximize the likehood of the target tokens conditioned on the image representations $I$ and previously generated tokens $y_{1:j-1}$,
  \vspace{-0.2cm}
  \begin{equation}
	Loss_{obj}=-\sum_{j=1}^{N}w_{j}\text{log}P(\hat{m}_{j}|I,m_{1:j-1}),
  \label{eq:eq1}
  \vspace{-0.2cm}
  \end{equation}
  where $m$ and $\hat{m}$ denote the input and target sequence, respectively. $N$ is the length of sequence. $w_{j}$ represents the weight for the $j$-th token, 0 is assigned to the format transcription token and 1 is assigned to other tokens to ensure the model is trained to predict the desired tokens but not the format transcription tokens. 
  
  \textbf{Inference Pipeline}. We take the format transcription token as the prompt to achieve the format-specific detection and the sequence decoder generates the rest of the sequence conditioned on the prompt and image representation. Once the whole sequence is generated, we de-quantize the sequence to obtain the location of lane.
  More details about de-quantization for each format can be found in the supplementary materials. Inference process is presented in Fig.~\ref{fig:short_2}.
  
  \subsection{Multi-Format Model Tuning Based on Reinforcement Learning}
  \label{sec:tuning}
  Lane2Seq get rids of the complicated task-specific components, such as task-specific head networks, via the sequence generation. However, this task-agnostic architecture inevitably lacks the task-specific knowledge of the lane detection, making the model ineffective in learning the features of lane. Exising lane detection methods~\cite{zheng2021resa,honda2023clrernet} usually design the specific modules to incorporate the task-specific knowledge into the model.
  But this way makes the model architecture complicate. We propose a novel Multi-Format model tuning method based on Reinforcement Learning (MFRL) to learn task-specific knowledge effectively without changing any components of the model.
  
  Inspired by Task-Reward~\cite{pinto2023tuning}, MFRL takes the evaluation metric, which naturally contains the task-specific knowledge, as the reward function and adopts the reinforcement learning to tune the model, which is common practice in the modern language model~\cite{brown2020language,du2023guiding}. Specifically, 
  it is difficult for the model to converge if model is trained with reinforcement learning method from scratch~\cite{pinto2023tuning}. Hence, MFRL consists of two stages: the pretraining stage and the model tuning stage. 
  In the pretraining stage, model is trained on the lane detection dataset with objective function in Eq.~\ref{eq:eq1} to have a good weight initialization.
  
  In the model tuning stage, we use the REINFORCE algorithm~\cite{sutton1999policy} to maximize the objective function as below,
  \begin{equation}
	Obj=E_{c \sim D}\left[E_{t \sim Q(\cdot|c,u)}R(t,g)\right],
	\vspace{-0.2cm}
  \label{eq:eq10}
  \end{equation}
  where $c$, $t$, and $g$ represent the input image, generated format-specific sequences, and ground truths, respectively. $E$ and $u$ denote the mathematic expectation and the model parameter. $R$, $D$, and $Q$ stand for the reward function, data distribution of the dataset, and conditional distribution parameterized by $u$.
  REINFORCE algorithm estimates the gradient of the reward function by
  \begin{equation}
	\triangledown_{u} E_{t \sim Q}[R(t,g)] = E_{t \sim Q}[R(t,g)\triangledown_{u}\text{log}Q(t|c,u)],
  \label{eq:eq2}
  \end{equation}
  In practice, Eq.~\ref{eq:eq2} is computed as an average of per-sequence gradient and reward function of per-sequence is not required to be 
  differentiable. More details about REINFORCE algorithm can be found in the supplementary materials.
  
  \textbf{Reward Function}. Evaluation metric F1 score and accuracy are unable to adopted as the reward directly, because thay cannot be decomposable as a sum of per-sequence rewards (see Sec~\ref{sec:4.2}). However, F1 score or accuracy is composed of false positives ($FP$), true positives ($TP$), and false negatives ($FN$). All three indicators can be computed per-sequence.
  We opt to use $TP$ and $FP$ to design the reward. Specifically, reward for three detection formats are constructed as follows.
  
  Reward for the \textit{segmentation} format (termed as $R_{seg}$). Segmentation-based lane detection includes two task-specific knowledge, i.e., knowledge of the segmentation and lane detection. For the knowledge of lane detection, we adopt a matched Line-IOU (LIOU)~\cite{zheng2022clrnet}, which can demonstrate the quality of location prediction and shape prediction of the lane. For the knowledge of the segmentation, we adopt the matched Mean Intersection over Union (mIOU), which is a widely
  used metric in the semantic segmentation to evaluate the quality of segmentation, as the reward. Finally, we add both of them to compute $r_{seg}$ as below,
  \begin{equation}
  \small
  \begin{aligned}
	R_{seg}(t,g)=&\frac{1}{K}\sum_{k=1}^{K}[LIOU(p_{k},g)+mIOU(p_{k},g)] \\
	&- \lambda_{1} FP_{seg}(t,g),
  \end{aligned}
	\vspace{-0.1cm}
	\label{eq:eq3}
  \end{equation}
  where $K$ and $p_{k}$ represents the number of true positives in $t$ and the $k$-th true positive. $FP_{seg}$ is the false positive rate of the segmentation format and $\lambda_{1}$ is the weight to control the affect of $FP_{seg}$. We introduce $FP_{seg}$ to penalize for false positives.
  
  Reward for the \textit{anchor} format (termed as $R_{a}$). Anchor-based lane detection contains the knowledge of the keypoint location and lane detection. For the knowledge of keypoint location, we simply adopt the matched Euclidean distance $d(p_{k},g)$ between true positives and corresponding ground truths as the reward. Since MFRL needs to maximize the reward, we rescale the reward as $d_{r}(p_{k},g)=1-\frac{d(p_{k},g)}{H}$, where $H$ is the height of image.
  For the knowledge of lane detection, we also utilize the matched LIOU as the reward. The formulation of $r_{a}$ is:
  \begin{equation}
	\small
	  R_{a}(t,g)=\frac{1}{K}\sum_{k=1}^{K}[LIOU(p_{k},g)+d_{r}(p_{k},g)]- \lambda_{2} FP_{a}(t,g),
	\label{eq:eq4}
	\end{equation}
  where $FP_{a}$ and $\lambda_{2}$ stand for the false positives rate of the anchor format and the weight of $FP_{a}$.
  
  Reward for the \textit{parameter} format (termed as $R_{p}$). Parameter-based detection only has the knowledge of lane detection, hence we directly use the matched LIOU as the reward.
  \vspace{-0.1cm}  
  \begin{equation}
	\small
	  R_{p}(t,g)=\frac{1}{K}\sum_{k=1}^{K}LIOU(p_{k},g)- \lambda_{3} FP_{p}(t,g),
	\label{eq:eq5}
	\vspace{-0.2cm} 
	\end{equation}
  where $FP_{p}$ and $\lambda_{3}$ stand for the false positives rate of the parameter format and the weight of $FP_{p}$. 
  
  Since different detection formats have different contributions, we weight the objective function of different formats. The final objective function of MFRL can be computed as,
  \begin{equation}
	\small
	  Obj_{total} = \lambda_{4}Obj_{seg}+\lambda_{5}Obj_{a}+\lambda_{6}Obj_{p},
	\label{eq:eq7}
	\end{equation}
  where $Obj_{seg}$, $Obj_{a}$, and $Obj_{p}$ denote the objective function of segmentation, anchor, and parameter format, respectively. The objective function of three formats is the same as Eq.~\ref{eq:eq10}, except the format-specific sequence and reward function. $\lambda_{4}$, $\lambda_{5}$, and $\lambda_{6}$ are the scale factor.
  
  \begin{table*}[!t]
	\centering
	\caption{Comparison of F1 score and MACs (multiply–accumulate operations) on CULane testing set. We only report the false positives for “Cross” category following~\cite{zheng2022clrnet}.}
	\tabcolsep=0.5cm
	\Huge
	\resizebox{\textwidth}{!}{
	\begin{tabular}{ccccccccccccc}
	  \toprule
	  Methods& Encoder& Normal$\uparrow$ & Crowded$\uparrow$& Dazzle$\uparrow$& Shadow$\uparrow$& No line$\uparrow$& Arrow$\uparrow$& Curve$\uparrow$& Night$\uparrow$& Cross$\downarrow$& Total$\uparrow$& MACs(G)$\downarrow$ \\
	  \midrule
	  \bf Segmentation based &&&&&&&&&&&& \\
	  \cline{1-1}
	  SCNN~\cite{pan2018spatial}&ResNet50& 90.60& 69.70& 58.50& 66.90& 43.40& 84.10& 64.40& 66.10& 1900& 71.60& 8.4 \\
	  RESA~\cite{zheng2021resa}&ResNet50& 92.10& 73.10& 69.20& 72.80& 47.70& 88.30& 70.30& 69.90& 1503& 75.3& 7.6 \\
	  AtrousFormer~\cite{yang2023lane}&ResNet34& 92.83& 75.96& 69.48& 77.86& 50.15& 88.66& 71.14& 73.74& \bf 1054& 78.08& 14.22 \\
	  LaneAF~\cite{yang2023lane}&DLA34~\cite{yu2018deep}& 91.80& 75.61& 71.78& 79.12& 51.38& 86.88& 72.70& 73.03& 1360& 77.41& 12.73 \\
	  \rowcolor{gray!25}
	  Lane2Seq (segmentation)&ViT-Base& \bf 93.39& \bf 77.27& \bf 73.45& \bf 79.69& \bf 53.91& \bf 90.53& \bf 73.37& \bf 74.96& 1129& \bf 79.64& 15.19 \\
	  \cline{1-1}
	  \bf Anchor based &&&&&&&&&&&& \\
	  \cline{1-1}
	  LaneATT~\cite{tabelini2021keep}&ResNet122& 91.74& 76.16& 69.47& 76.31& 50.46& 86.29& 64.05& 70.81& 1264& 77.02& 70.5 \\
	  O2SFormer~\cite{zhou2023end}&ResNet50& 93.09& 76.57& 72.25& 76.56& 52.80& 89.50& 69.60& 73.85& 3118& 77.83& 27.51 \\
	  UFLD~\cite{qin2020ultra}&ResNet34& 90.70& 70.20& 59.50& 69.30& 44.40& 85.70& 69.50& 66.70& 2037& 72.30& 3.3 \\
	  CondLaneNet~\cite{liu2021condlanenet}&ResNet34& 93.38& 77.14& 71.17& \bf 79.93& 51.85& 89.89& \bf 73.88& 73.92& 1387& 78.74& 19.6 \\
	  ADNet~\cite{xiao2023adnet}&ResNet34& 92.90& 77.45& 71.71& 79.11& 52.89& 89.90& 70.64& 74.78& 1499& 78.94& 18.6 \\
	  CLRNet~\cite{zheng2022clrnet}&ResNet34& \bf 93.49& \bf 78.06& \bf 74.57& 79.92& \bf 54.01& \bf 90.59& 72.77& 75.02& 1216& {\bf 79.73}& 17.3\\
	  \rowcolor{gray!25}
	  Lane2Seq (anchor)&ViT-Base& 93.11& 77.43& 73.25& 79.46& 53.74& 90.02& 72.44& \bf 75.12& \bf 1173& 79.27& 15.19 \\
	  \cline{1-1}
	  \bf Parameter based &&&&&&&&&&&& \\
	  \cline{1-1}
	  LSTR~\cite{liu2021end}&ResNet18& -& -& -& -& -& -& -& -& -& 64.00& 5.9 \\
	  BezierLaneNet~\cite{feng2022rethinking}&ResNet18& 90.22& 71.55& 62.49& 70.91& 45.30& 84.09& 58.98& 68.70& 996& 73.67& 4.7 \\
	  BSNet~\cite{chen2023bsnet}&ResNet34& \bf 93.75& \bf 78.01& \bf 76.65& \bf 79.55& \bf 54.69& \bf 90.72& \bf 73.99& \bf 75.28& 1445& \bf 79.89& 17.2 \\
	  Eigenlanes~\cite{jin2022eigenlanes}&ResNet50& 91.70& 76.00& 69.80& 74.10& 52.20& 87.70& 62.90& 71.80& 1509& 77.20& 18.7 \\
	  Laneformer~\cite{han2022laneformer}&ResNet50&91.77& 75.74& 70.17& 75.75& 48.73& 87.65& 66.33& 71.04& {\bf 19}& 77.06& 26.2 \\
	  \rowcolor{gray!25}
	  Lane2Seq (parameter)&ViT-Base& 93.03& 76.42& 72.17& 78.32& 52.89& 89.67& 72.67& 73.98& 1319& 78.39& 15.19 \\
	  \bottomrule
	\end{tabular}}
	\label{sec:table}
  \end{table*}
  
  \section{Experiments}
  \subsection{Datasets}
  We conduct experiments on three lane detection benchmarks: CULane~\cite{pan2018spatial}, Tusimple~\cite{tusimple}, and LLAMAS~\cite{behrendt2019unsupervised}.
  
  {\bf CULane} is a widely used large-scale dataset for lane detection. 
  It contains a lot of challenging scenarios such as crowded roads. The CULane dataset 
  consists of 88.9K images for training, 9.7K images in the validation set, and 34.7K images for the test. Image size is 1640×590. 
  
  {\bf Tusimple} is a real highway dataset consisting of 3,626 training images and 2,782 testing images. All images have 1280×720 pixels.
  
  {\bf LLAMAS} is a recently released large-scale lane detection dataset with over 100K images. All lane markers are annotated with high-accurate maps. Image size is 1280×717.
  
  \subsection{Evaluation Metrics}
  \label{sec:4.2}
  For CULane and LLAMAS dataset, we adopt the F1 score to measure the performance: $F_1=\frac{2\times P r e c i s i o n\times R e c a l l}{Precision+Recall}$, where $Precision=\frac{TP}{TP+FP}\ $ and $Recall=\frac{TP}{TP+FN}$. 
  
  For Tusimple dataset, we use F1 score, accuracy, false positives, and false negatives to evaluate the model performance. Accuracy is defined as $Accuracy=\frac{\sum_{c l i p} C_{c l i p}}{\sum_{c l i p} S_{c l i p}}$,
  where $C_{c l i p}$ represents the number of accurately predicted lane points and $S_{c l i p}$ denotes the total number of lane points of a clip.
  A lane point is considered as correct if its distance is smaller than the given threshold $t_{pc}=\frac{20}{cos(a_{yl})}$, where $a_{yl}$ denotes the angle of the corresponding ground truth.
  
  \subsection{Implementation Details}
  \textbf{Model}. We adopt the ViT-Base initialized with the MAE~\cite{he2022masked} pretrained parameters as the image encoder. The patch size is 16×16. The decoder is composed of 2 transformer blocks and its hidden size is 256. The number of attention heads is 8 and the hidden size of feed-forward network is 1024.
  $n_{bins}$ is set to 1000 and the dimension of word embedding of the vocabulary is 256. Our Lane2Seq is implemented based on MMDetection toolbox~\cite{mmdetection}. $\lambda_{1}$, $\lambda_{2}$, $\lambda_{3}$, $\lambda_{4}$, $\lambda_{5}$, and $\lambda_{6}$ are 
  set to 0.3, 0.3, 0.1, 0.2, 1, and 1.5, respectively. 
  
  \textbf{Training}. All input images are resized to 320×800. We utilize the AdamW~\cite{loshchilov2017decoupled} as the optimizer with initial learning rate 1e-4.
  The training epochs of the pretraining stage are 5, 20, and 15 for CULane, Tusimple, and LLAMAS. We set the training epochs of the model tuning stage to 15, 30, and 55 for CULane, Tusimple, and LLAMAS, respectively.
  For the data augmentation, we use the random horizontal flips and random affine transformation including scaling, rotation, and translation. All experiments are conducted on 8 A100 GPUs with total batch size 384.

  \subsection{Comparison with the State-of-the-art Methods}
  \textbf{Performance on LLAMAS}. Our Lane2Seq achieves a new state-of-the-art performance on LLAMAS dataset using segmentation format. In Table~\ref{sec:table2}, Lane2Seq increases F1 score from 97.16\% to 97.42\% compare to the previous state-of-the-art method CLRNet. 
  Compared with parameter-based state-of-the-art method BezierLaneNet, Lane2Seq performs better than BezierLaneNet (96.74\% vs. 96.11\%). The results manifest the detection strengths of Lane2Seq in the multi-lane scenario (The number of lanes $\geq$ 5), such as highways.
  The reason may be that transformer-based architecture has the advantages in capturing long-range dependency.
  
  \begin{table}[!t]
	\centering
	\caption{Comparison of the performance of different models on LLAMAS.}
	\tabcolsep=0.5cm
	\resizebox{\linewidth}{!}{
	\begin{tabular}{cccc}
	  \toprule
	  Methods& Encoder & F1(\%)$\uparrow$ &  \\
	  \midrule
	  PolyLaneNet~\cite{tabelini2021polylanenet}&EfficientNet-b0~\cite{tan2019efficientnet}& 90.20 \\
	  BezierLaneNet & ResNet34 & 96.11 \\
	  LaneATT & ResNet34 & 94.96 \\
	  LaneATT & ResNet122 & 95.17 \\
	  LaneAF & DLA34 & 96.90 \\
	  CLRNet & ResNet18 & 96.96 \\
	  CLRNet & DLA34 & 97.16 \\
	  \midrule
	  \rowcolor{gray!25}
	  Lane2Seq (segmentation) & ViT-Base & \bf 97.42 \\
	  \rowcolor{gray!25}
	  Lane2Seq (anchor) & ViT-Base & 97.05 \\
	  \rowcolor{gray!25}
	  Lane2Seq (parameter) & ViT-Base & 96.74 \\
	  \bottomrule
	\end{tabular}}
	\vspace{-0.4cm}
	\label{sec:table2}
  \end{table}
  
  \begin{table}[!t]
	\centering
	\caption{Comparison of the performance of different models on Tusimple. Acc denotes accuracy.}
	\tabcolsep=0.5cm
	\Huge
	\resizebox{\linewidth}{!}{
	\begin{tabular}{ccccccc}
	  \toprule
	  Methods& Encoder & F1(\%)$\uparrow$ & Acc(\%)$\uparrow$ & FP(\%)$\downarrow$ & FN(\%)$\downarrow$ &\\
	  \midrule
	  SCNN&VGG16& 95.97 & 96.53 & 6.17 & \bf 1.80 \\
	  RESA&ResNet34& 96.93 & 96.82 & 3.63 & 2.48 \\
	  PolyLaneNet&EfficientNet-bo& 90.62 & 93.36 & 9.42 & 9.33 \\
	  UFLD&ResNet34& 88.02 & 95.86 & 18.91 & 3.75 \\
	  LaneATT&ResNet122& 96.06 & 96.10 & 5.64 & 2.17 \\
	  GANet~\cite{wang2022keypoint}&ResNet34& 97.68 & 95.87 & \bf 1.99 & 2.64 \\
	  CondLaneNet&ResNet34& 96.98 & 95.37 & 2.20 & 3.82 \\
	  CondLaneNet&ResNet101& 97.24 & 96.54 & 2.01 & 3.50 \\
	  CLRNet&ResNet34& 97.82 & \bf 96.87 & 2.27 & 2.08 \\
	  CLRNet&ResNet101& 97.62 & 96.83 & 2.37 & 2.38 \\
	  \midrule
	  \rowcolor{gray!25}
	  Lane2Seq (segmentation) & ViT-Base & \bf 97.95 & 96.85 & 2.01 & 2.03 \\
	  \rowcolor{gray!25}
	  Lane2Seq (anchor) & ViT-Base & 97.86 & 96.72 & 2.21 & 2.05 \\
	  \rowcolor{gray!25}
	  Lane2Seq (parameter) & ViT-Base & 96.59 & 96.00 & 2.23 & 3.54 \\
	  \bottomrule
	\end{tabular}}
	\label{sec:table3}
  \end{table}
  
  \textbf{Performance on Tusimple}. Table~\ref{sec:table3} presents the performance comparsion with state-of-the-art approaches on Tusimple. Due to small data scale and single scene,
  the performance gap between different models is small. Our Lane2Seq using segmentation format sets a new state-of-the-art F1 score of 97.95\%. The results also validate the detection strengths of Lane2Seq
  in multi-lane scenario.
  
  \textbf{Performance on CULane}. We compare Lane2Seq with other state-of-the-art methods on CULane and results are shown in Table~\ref{sec:table}. For the segmentation-based methods, Lane2Seq improves F1 score from 78.08\% to 79.64\% compared to the previous state-of-the-art methods AtrousFormer.
  Compared with anchor-based methods, Lane2Seq outperforms most previous methods. For example, Lane2Seq achieves the better performance than existing row anchor-based methods, such as CondLaneNet (79.27\% vs. 78.74\%). Moreover, Lane2Seq achieves competitive performance compared to CLRNet (79.27\% vs. 79.73\%).
  One possible reason for the performance gap between CLRNet and Lane2Seq is that CLRNet adopts multi-scale detection, which can notably improve the detection performance, while Lane2Seq only uses the plain transformer architecture with single-scale detection. For the comparsion of parameter-based methods,
  Lane2Seq surpasses the existing transformer-based methods. For example, Lane2Seq significantly increases F1 score from 64.00\% to 78.39\% compared with LSTR and outperforms Laneformer by 1.33\% (78.39\% vs. 77.06\%). But Lane2Seq is ranked second after BSNet. 
  
  Based on the above results and analysises, we can conclude that the unified lane detection via the sequence generation without any well-designed task-specific components, combining our reinforcement learning-based multi-format model tuning, can achieve promising performance.
  
\begin{figure*}[t!]
	\centering
	\includegraphics[width=10cm]{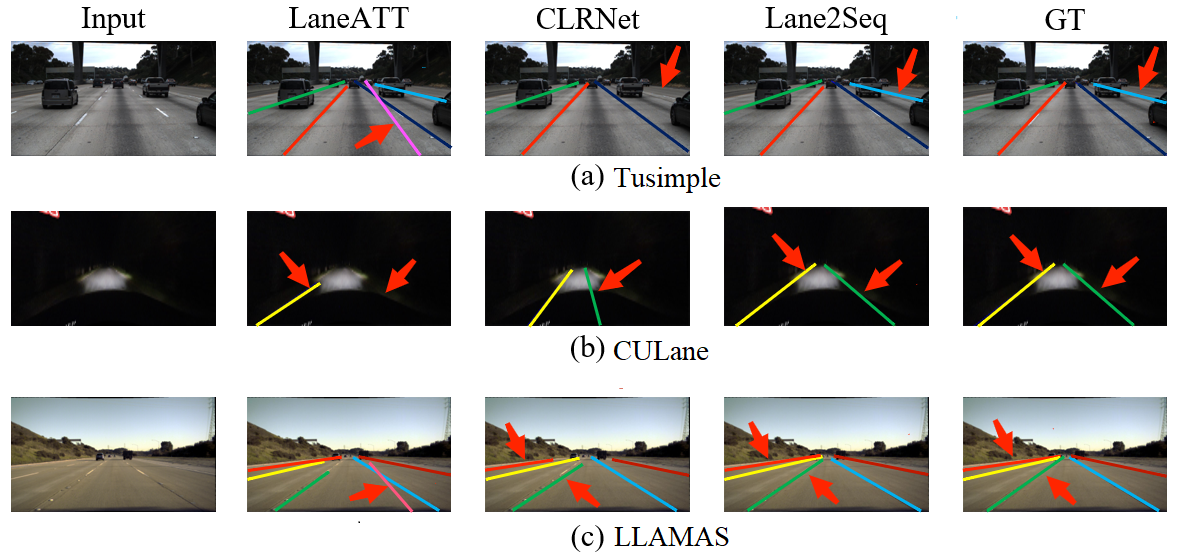}
	\caption{Visualization results of LaneATT, CLRNet, and Lane2Seq on Tusimple, CULane, and LLAMAS. }
	\label{fig:short4}
	\vspace{-0.3cm}
	\end{figure*}

  \textbf{Qualitative Results}. We display the qualitative results in Fig.~\ref{fig:short4}. The results show that Lane2Seq can effectively detect lanes in the multi-lane scenario (see Fig~\ref{fig:short4} (a) and (c)).
  Even in the case of the night scene, Lane2Seq successfully discriminates the lanes (see Fig~\ref{fig:short4} (b)). 
  
  \subsection{Ablation Study}
  We conduct the ablation experiments on CULane dataset to validate the effectiveness of each component. More ablation studies are in the supplementary materials.
  
  \textbf{Multi-Format Model Tuning Based on Reinforcement Learning}. We first ablate the effectiveness of the proposed Multi-Format Model Tuning Based on Reinforcement Learning (MFRL). 
  As shown in Table~\ref{sec:table4}, MFRL significantly improves the performance of different detection formats. For the segmentation format, MFRL gains 15.37\% F1 score improvement (79.64\% vs. 64.27\%).
  For the anchor and parameter format, MFRL increases 13.04\% (79.27\% vs. 66.23\%) and 10.25\% (78.39\% vs. 68.14\%) F1 score, respectively. Reasons of performance improvement can be attributed in two folds: (1) the vanilla Lane2Seq
  is a task-agnostic architecture and lacks the task-specific knowledge, leading to the ineffective feature learning of lane. MFRL tackles this problem by adopting the evaluation metric, which is designed based on the lane prior information like angle, as the reward
  to tune the model. (2) MFRL makes the model's predictions aligned with their intended usage, i.e., how to achieve a high performance. Optimizing the loss function is a common practice in the compute vision. 
  However, this way indirectly optimizes for model's intended usage, because the lower loss does not mean the higher performance. Instead, MFRL views the maximizing the evaluation metric as the optimization objective, which is positively correlated with the model performance.
  
  \begin{table}[!t]
	\centering
	\caption{Ablation study on the multi-format model tuning based on reinforcement learning.}
	\tabcolsep=0.5cm
	\Huge
	\resizebox{\linewidth}{!}{
	\begin{tabular}{cccccc}
	  \toprule
	  Methods& F1(\%)$\uparrow$ &Precision(\%)$\uparrow$ &Recall(\%)$\uparrow$ &  \\
	  \midrule\
	  Lane2Seq (segmentation) & 64.27 & 70.44 & 52.77 \\
	  \rowcolor{gray!25}
	  Lane2Seq (segmentation) + MFRL & \bf 79.64 (+\text{\color{red} 15.37}) & \bf 85.26 (+\text{\color{red} 14.82}) & \bf 69.00 (+\text{\color{red} 16.23}) \\
	  \midrule
	  Lane2Seq (anchor) & 66.23 & 71.36 & 55.75 \\
	  \rowcolor{gray!25}
	  Lane2Seq (anchor) + MFRL & \bf 79.27 (+\text{\color{red} 13.04}) & \bf 84.72 (+\text{\color{red} 13.36}) & \bf 67.28 (+\text{\color{red} 11.53}) \\    
	  \midrule
	  Lane2Seq (parameter) & 68.14 & 73.27 & 57.89 \\
	  \rowcolor{gray!25}
	  Lane2Seq (parameter) + MFRL & \bf 78.39 (+\text{\color{red} 10.25}) & \bf 83.68 (+\text{\color{red} 10.41}) & \bf 66.94 (+\text{\color{red} 9.05}) \\    
	  \bottomrule
	\end{tabular}}
	\label{sec:table4}
  \end{table}
  
  \begin{table}[!t]
	\centering
	\caption{Ablation study on the different reward functions. F1 score is adopted as the indicator.}
	\tabcolsep=0.5cm
	\Huge
	\resizebox{\linewidth}{!}{
	\begin{tabular}{cccccc}
	  \toprule
	  $R_{seg}$& $R_{a}$& $R_{p}$& Segmentation& Anchor& Parameter \\
	  \midrule
	  &&&64.27 & 66.23 & 68.14 \\ 
	  \midrule
	  \checkmark&&&77.69 (+\text{\color{red} 13.42}) & 66.23 (+\text{\color{blue} 0.0}) & 68.14 (+\text{\color{blue} 0.0})\\ 
	  &\checkmark&&64.27 (+\text{\color{blue} 0.0}) & 76.94 (+\text{\color{red} 10.71}) & 68.14 (+\text{\color{blue} 0.0})\\
	  &&\checkmark&64.27 (+\text{\color{blue} 0.0})& 66.23 (+\text{\color{blue} 0.0}) & 77.02 (+\text{\color{red} 8.88})\\
	  \checkmark&\checkmark&&78.72 (+\text{\color{red} 14.45}) & 78.14 (+\text{\color{red} 11.91})& 68.14 (+\text{\color{blue} 0.0})\\
	  \checkmark&&\checkmark&78.70 (+\text{\color{red} 14.43}) & 66.23 (+\text{\color{blue} 0.0})& 77.90 (+\text{\color{red} 9.76})\\
	  &\checkmark&\checkmark&64.27 (+\text{\color{blue} 0.0}) & 78.20 (+\text{\color{red} 11.97})& 77.93 (+\text{\color{red} 9.79})\\
	  \checkmark&\checkmark&\checkmark& \bf 79.64 (+\text{\color{red} 15.37}) & \bf 79.27 (+\text{\color{red} 13.04})& \bf 78.39 (+\text{\color{red} 10.25})\\
	  \bottomrule
	\end{tabular}}
	\vspace{-0.2cm}
	\label{sec:table5}
  \end{table}
  
  \textbf{Effectiveness of Reward Function of Different Formats}. We ablate the effectiveness of reward function of different formats and results are shwon in Table~\ref{sec:table5}. It can be observed that our reward function can bring consistent performance improvement.
  However, we can also see that the performance will not be improved if the corresponding reward is not optimized. Besides, we can find that multi-reward optimization can achieves better results than single-reward optimization. Reason may be that multi-reward optimization enables model to learn correlations between different detection formats and provides more supervised signals.
  
  \textbf{Comparison with other Multi-Task Methods}. An alternative method to unify different detection formats is using different detection heads in a model. YOLOP~\cite{wu2022yolop} is a representative work for this method.
  We replace the heads of YOLOP with anchor-based head, segmentation-based head, and parameter-based head. The segmentation-based loss, anchor-based loss, parameter-based loss for YOLOP are same as that in SCNN~\cite{pan2018spatial}, LaneATT~\cite{tabelini2021keep}, and PolyLaneNet~\cite{tabelini2021polylanenet}.
  From Table~\ref{sec:table6}, we can observe that Lane2Seq surpasses YOLOP for all detection formats. It should be noted that YOLOP contains task-specific modules, such as detection head and loss function, but Lane2Seq does not.
  Results manifest the sequence generation combining with MFRL is a more simple and effective way to unify lane detection.
  
  \begin{table}[!t]
	\centering
	\caption{Comparison between Lane2Seq and YOLOP.}
	\tabcolsep=0.5cm
	\Huge
	\resizebox{\linewidth}{!}{
	\begin{tabular}{cccccc}
	  \toprule
	  Methods& F1(\%)$\uparrow$ &Precision(\%)$\uparrow$ &Recall(\%)$\uparrow$ &  \\
	  \midrule
	  YOLOP (segmentation)  & 75.23 & 79.46 & 60.58 \\
	  \rowcolor{gray!25}
	  Lane2Seq (segmentation) & \bf 79.64 (+\text{\color{red} 4.41}) & \bf 85.26 (+\text{\color{red} 5.8}) & \bf 69.00 (+\text{\color{red} 8.42}) \\
	  \midrule
	  YOLOP (anchor)  & 76.44 & 80.99 & 62.15 \\
	  \rowcolor{gray!25}
	  Lane2Seq (anchor) & \bf 79.27 (+\text{\color{red} 2.83}) & \bf 84.72 (+\text{\color{red} 3.73}) & \bf 67.28 (+\text{\color{red} 5.13}) \\    
	  \midrule
	  YOLOP (parameter)  & 69.43 & 74.38 & 58.46 \\
	  \rowcolor{gray!25}
	  Lane2Seq (parameter) & \bf 78.39 (+\text{\color{red} 8.96}) & \bf 83.68 (+\text{\color{red} 9.30}) & \bf 66.94 (+\text{\color{red} 8.48}) \\   
	  \bottomrule
	\end{tabular}}
	\vspace{-0.5cm}
	\label{sec:table6}
  \end{table}
  
  \textbf{Ablation Study on the Scale Factor of Different Objective Functions}. For the limitation of the space, we take the segmentation format as an example to present the influence of the scale factor of different objective functions in MFRL. Results of other two formats can be found in the supplementary materials.
  As shown in Table~\ref{sec:table7}, assigning the same factor to different objective functions causes insuperior performance, indicating different detection formats have different contributions. We set $\lambda_{4}$, $\lambda_{5}$, and $\lambda_{6}$ to 0.2, 1, and 1.5 according to the model performance.
  
  \begin{table}[!t]
	\centering
	\caption{Ablation study on different scale factors.}
	\tabcolsep=0.5cm
	\centering
	\tiny
	\resizebox{\linewidth}{!}{
	\begin{tabular}{cccc}
	  \toprule
	  $\lambda_{4}$ & $\lambda_{5}$ & $\lambda_{6}$ & F1(\%)$\uparrow$ \\
	  \midrule
	  1 & 1 & 1 & 73.23 \\
	  \midrule
	  0.7 & 1 & 1 & 74.96 \\
	  0.4 & 1 & 1 & 76.27 \\
	  0.2 & 1 & 1 & 77.35 \\
	  0.1 & 1 & 1 & 77.02 \\
	  \midrule
	  0.2 & 1.1 & 1 & 77.03 \\
	  0.2 & 1.2 & 1 & 76.75 \\
	  0.2 & 0.9 & 1 & 77.09 \\
	  0.2 & 0.8 & 1 & 77.00 \\
	  \midrule
	  0.2 & 1 & 0.9 & 76.87 \\
	  0.2 & 1 & 1.2 & 77.37 \\
	  0.2 & 1 & 1.4 & 79.46 \\
	  \rowcolor{gray!25}
	  0.2 & 1 & 1.5 & \bf 79.64 \\
	  0.2 & 1 & 1.6 & 79.39 \\
	  \bottomrule
	\end{tabular}}
	\vspace{-0.3cm}
	\label{sec:table7}
  \end{table}
  
  \section{Conclusion}
  This paper presents a novel sequence generation-based lane detection framework, i.e., Lane2Seq, to unify lane detection, which casts lane detection as a sequence generation task. 
  Lane2Seq gets rid of complicated task-specific modules and adopts a simple transformer-based encoder-decoder architecture. To incorporate the task-specific knowledge, we employ a 
  multi-format model tuning based on reinforcement learning (MFRL). Extensive experiments show Lane2Seq is effective and achieves competitive results compared to the state-of-the-art methods.
  
  \textit{Limitation}. One major limitation of Lane2Seq is that sequence generation model is expensive for long sequence (mainly for inference). The inference speed is low when there are more than 5 lanes in an image
  despite its detection strength in multi-lane scenario. Therefore, future work is required to make it faster for real-time lane detection applications. Another limitation is that MFRL can only be applied to sequence generation-based model 
  currently. Our ongoing work is applying MFRL to other vision models.

\clearpage
{
    \small
    \bibliographystyle{ieee_fullname}
    \bibliography{main}
}
\clearpage
\appendix

\section{De-quantization for different formats}
Different detection formats require a specific de-quantization scheme to obtain the final predictions. A detailed description of de-quantization for each detection format is given below.

De-quantization for the \textit{segmentation} format. we dequantize the coordinates tokens corresponding to each polygon and then convert them into the mask. Specifically, given the x,y coordinates, their dequantization process can be expressed by $x = \frac{x}{n_{bins}} \times Wid, y = \frac{y}{n_{bins}} \times Hei$,
where $Wid$ and $Hei$ represent the width and height of the image, respectively.

De-quantization for the \textit{anchor} format. We directly dequantize the image coordinate tokens of the keypoints, whose dequantization process is the same as that of a polygon.

De-quantization for the \textit{parameter} format. We dequantize the parameter tokens and vertical offset corresponding to each parameter sequence. The dequantization of parameter token can be expressed by $a_{i} = \frac{a_{i}}{n_{bins}} \times desigmoid(a_{i})$, where $desigmoid$ is the inverse function of the sigmoid.
The dequantization of vertical offset is $s = \frac{s}{n_{bins}} \times Hei$.

\begin{figure}[!t]
  \begin{center}
  \includegraphics[width=\linewidth]{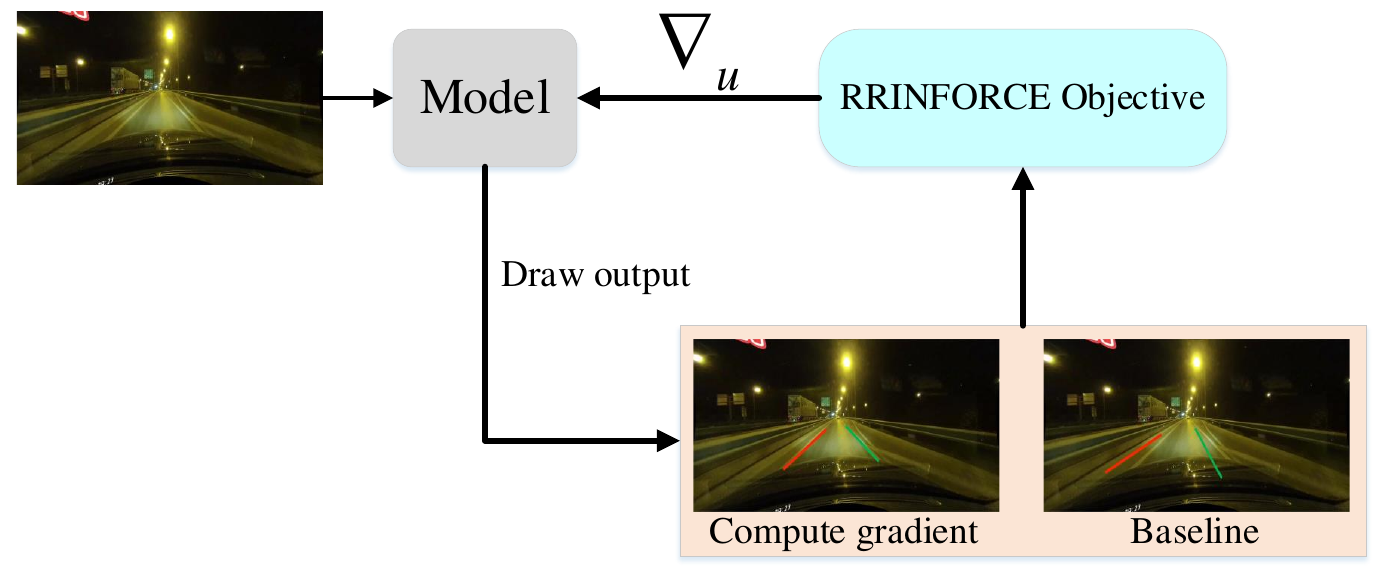}
  \end{center}
     \caption{Illustration of REINFORCE algorithm.}
  \label{fig:al_2}
  \end{figure}

\section{REINFORCE algorithm}
REINFORCE algorithm~\cite{williams1992simple} is a widely used method to maximize the reward function in reinforcement learning. Given an input image $c$,
REINFORCE algorithm estimates the gradient of the reward function as below,
\begin{equation}
  \triangledown_{u} E_{t \sim Q}[R(t,g)] = E_{t \sim Q}[R(t,g)\triangledown_{u}\text{log}Q(t|c,u)],
\label{eq:ai_1}
\end{equation}
where $t$, and $g$ represent the generated format-specific sequences, and ground truths, respectively. $E$ and $u$ denote the mathematic expectation and the model parameter. $R$, $D$, and $Q$ stand for the reward function, data distribution of the dataset, and conditional distribution parameterized by $u$.
In order to reduce the variance of the gradient estimate, REINFORCE algorithm usually subtracts a baseline value $b$ from the reward function. As presented in Fig.~\ref{fig:al_2}, REINFORCE first draws two outputs from one training image, using one to estimate the gradient and the other to compute the baseline value.
The procedure of REINFORCE algorithm is: (1) Draw two outputs from one input image. (2) Compute the reward function $R_{gradient}$ and $R_{baseline}$, whose formulation are the same as $R$. Final reward $r$ is compute by $r=R_{gradient}-R_{baseline}$. (3) Estimate the gradient according to Eq.~\ref{eq:ai_1} and $r$.

\section{Additional ablation studies}
We conduct additional ablation experiments on the hyper-parameters. If not specified, we still carry out experiments on CULane dataset.

\begin{figure}[!t]
  \begin{center}
  \includegraphics[width=\linewidth]{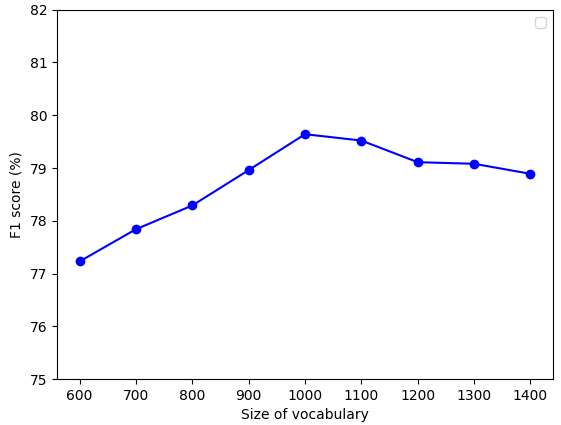}
  \end{center}
     \caption{Influence of the size of vocabulary.}
  \label{fig:al_3}
  \end{figure}

{\bf The size of vocabulary}. We first ablate the influence of the size of vocabulary $n_{bins}$ and the results are shown in Fig.~\ref{fig:al_3}. We take the segmentation format to conduct this experiment. Increasing the size of vocabulary $n_{bins}$ can improve the model performance because the quantization error is reduced accordingly. 
The performance declines when $n_{bins}$ is larger than 1000, thus we set $n_{bins}$ to 1000.

\begin{table}[!t]
  \centering
  \caption{Ablation study on different scale factors for anchor format.}
  \tabcolsep=0.5cm
  \centering
  \tiny
  \resizebox{\linewidth}{!}{
  \begin{tabular}{cccc}
    \toprule
    $\lambda_{4}$ & $\lambda_{5}$ & $\lambda_{6}$ & F1(\%)$\uparrow$ \\
    \midrule
    1 & 1 & 1 & 73.47 \\
    \midrule
    0.7 & 1 & 1 & 74.85 \\
    0.4 & 1 & 1 & 76.02 \\
    0.2 & 1 & 1 & 77.44 \\
    0.1 & 1 & 1 & 77.58 \\
    \midrule
    0.2 & 1.1 & 1 & 77.58 \\
    0.2 & 1.2 & 1 & 77.00 \\
    0.2 & 0.9 & 1 & 77.36 \\
    0.2 & 0.8 & 1 & 77.63 \\
    \midrule
    0.2 & 1 & 0.9 & 77.26 \\
    0.2 & 1 & 1.2 & 77.98 \\
    0.2 & 1 & 1.4 & 78.74 \\
    \rowcolor{gray!25}
    0.2 & 1 & 1.5 & \bf 79.27 \\
    0.2 & 1 & 1.6 & 79.00 \\
    \bottomrule
  \end{tabular}}
  \label{sec:table_1}
\end{table}

\begin{table}[!t]
  \centering
  \caption{Ablation study on different scale factors for parameter format.}
  \tabcolsep=0.5cm
  \centering
  \tiny
  \resizebox{\linewidth}{!}{
  \begin{tabular}{cccc}
    \toprule
    $\lambda_{4}$ & $\lambda_{5}$ & $\lambda_{6}$ & F1(\%)$\uparrow$ \\
    \midrule
    1 & 1 & 1 & 74.00 \\
    \midrule
    0.7 & 1 & 1 & 74.72 \\
    0.4 & 1 & 1 & 76.15 \\
    0.2 & 1 & 1 & 77.59 \\
    0.1 & 1 & 1 & 77.48 \\
    \midrule
    0.2 & 1.1 & 1 & 77.48 \\
    0.2 & 1.2 & 1 & 77.92 \\
    0.2 & 0.9 & 1 & 77.85 \\
    0.2 & 0.8 & 1 & 77.77 \\
    \midrule
    0.2 & 1 & 0.9 & 77.53 \\
    0.2 & 1 & 1.2 & 78.00 \\
    0.2 & 1 & 1.4 & 78.15 \\
    \rowcolor{gray!25}
    0.2 & 1 & 1.5 & \bf 78.39 \\
    0.2 & 1 & 1.6 & 77.97 \\
    \bottomrule
  \end{tabular}}
  \label{sec:table_2}
\end{table}

{\bf Additional ablation Study on the Scale Factor of Different Objective Functions}. We ablate the influence on the scale Factor of different objective functions for anchor and parameter format. Results are presented in Table~\ref{sec:table_1} and Table ~\ref{sec:table_2}. It can be seen that the model achieves the best performance when $\lambda_{4}$, $\lambda_{5}$, and $\lambda_{6}$ are 0.2, 1, and 1.5.

\begin{table}[!h]
  \centering
  \caption{Ablation study on the weight of false positives of segmentation format.}
  \tabcolsep=0.5cm
  \centering
  \resizebox{\linewidth}{!}{
  \begin{tabular}{cccc}
    \toprule
    $\lambda_{1}$ & F1(\%)$\uparrow$ & Precision(\%)$\uparrow$ & Recall(\%)$\uparrow$ \\
    \midrule
    0.0 & 77.59 & 83.79 & 67.85 \\
    0.1 & 78.06 & 84.11 & 67.98 \\
    0.2 & 78.89 & 85.00 & 68.87 \\
    0.3 & \bf 79.64 & \bf 85.26 & \bf 69.00 \\
    0.4 & 79.02 & 85.05 & 68.78 \\
    0.5 & 78.89 & 84.87 & 68.60 \\
    \bottomrule
  \end{tabular}}
  \label{sec:table_2_1}
\end{table}

{\bf Ablation study on the weight of false positives of segmentation format}. In Table~\ref{sec:table_2_1}, we ablate the influence of the weight of false positives of segmentation format $\lambda_{1}$. Performance increases when $\lambda_{1}$ increases, indicating introducing the penalty for false positives is beneficial.
However, performance declines when $\lambda_{1}$ is larger than 0.3, hence we set $\lambda_{1}$ to 0.3.

\begin{table}[!h]
  \centering
  \caption{Ablation study on the weight of false positives of anchor format.}
  \tabcolsep=0.5cm
  \centering
  \resizebox{\linewidth}{!}{
  \begin{tabular}{cccc}
    \toprule
    $\lambda_{2}$ & F1(\%)$\uparrow$ & Precision(\%)$\uparrow$ & Recall(\%)$\uparrow$ \\
    \midrule
    0.0 & 77.42 & 82.46 & 65.52 \\
    0.1 & 78.28 & 83.25 & 66.00 \\
    0.2 & 78.72 & 83.79 & 66.58 \\
    0.3 & \bf 79.27 & \bf 84.72 & \bf 67.28 \\
    0.4 & 79.10 & 84.15 & 66.95 \\
    0.5 & 78.75 & 83.89 & 66.58 \\
    \bottomrule
  \end{tabular}}
  \label{sec:table_3}
\end{table}

\begin{table}[!h]
  \centering
  \caption{Ablation study on the weight of false positives of parameter format.}
  \tabcolsep=0.5cm
  \centering
  \resizebox{\linewidth}{!}{
  \begin{tabular}{cccc}
    \toprule
    $\lambda_{3}$ & F1(\%)$\uparrow$ & Precision(\%)$\uparrow$ & Recall(\%)$\uparrow$ \\
    \midrule
    0.0 & 77.82 & 82.58 & 65.76 \\
    0.1 & \bf 78.39 & \bf 83.68 & \bf 66.94 \\
    0.2 & 78.21 & 83.60 & 66.38 \\
    0.3 & 77.89 & 83.11 & 66.02 \\
    0.4 & 77.56 & 82.26 & 65.28 \\
    0.5 & 77.12 & 81.83 & 64.87 \\
    \bottomrule
  \end{tabular}}
  \label{sec:table_4}
\end{table}

{\bf Ablation study on the weight of false positives of anchor format}. We ablate the influence of the weight of false positives of anchor format $\lambda_{2}$ and results are shown in Table~\ref{sec:table_3}. Similar to the performance trend of the segmentation format, model performance of the anchor format improves as the $\lambda_{2}$ increases. 
We set $\lambda_{2}$ to 0.3 according to the model performance.

{\bf Ablation study on the weight of false positives of parameter format}. We further ablate the influence of the weight of false positives of parameter format $\lambda_{3}$ and results are shown in Table~\ref{sec:table_4}. It can be seen that the weight of false positives of parameter format cannot be too large.
We set $\lambda_{3}$ to 0.1 according to the model performance.

\end{document}